\documentclass[11pt,letterpaper]{article}
\usepackage{emnlp2017}
\usepackage{times}
\usepackage{latexsym}

\usepackage{url}
\usepackage{mathtools}
\usepackage{amsmath}
\usepackage{amssymb}
\usepackage{bm}
\usepackage[ruled,linesnumbered]{algorithm2e}
\usepackage{multirow}
\usepackage[font={normalsize}]{caption}
\usepackage{subcaption}
\usepackage{graphicx}
\usepackage{esvect}
\usepackage{adjustbox}
\usepackage{hypcap}
\DeclareMathOperator*{\argmax}{arg\,max}

\usepackage{tikz}
\usetikzlibrary{arrows}
\usetikzlibrary{calc}
\usetikzlibrary{shapes.geometric}

\usetikzlibrary{positioning}
\usepackage{tkz-graph}

\usepackage{float}
\floatstyle{plaintop}
\restylefloat{table}

\usepackage[belowskip=-15pt,aboveskip=0pt]{caption}

\emnlpfinalcopy

\def\emnlppaperid{1420}

\title{DOC: Deep Open Classification of Text Documents}

\author{Lei Shu, Hu Xu, Bing Liu\\
    Department of Computer Science\\
    University of Illinois at Chicago\\
    \{lshu3, hxu48, liub\}@uic.edu
}

\date{}

\begin{document}
	\maketitle
	\begin{abstract}
		
		Traditional supervised learning makes the {\em closed-world} assumption that the classes appeared in the test data must have appeared in training. This also applies to text learning or text classification. As learning is used increasingly in dynamic open environments where some new/test documents may not belong to any of the training classes, identifying these novel documents during classification presents an important problem. This problem is called {\em open-world classification} or {\em open classification}. This paper proposes a novel deep learning based approach. It outperforms existing state-of-the-art techniques dramatically. 
		
		
	\end{abstract}
	
	\section{Introduction}
	
A key assumption made by classic supervised text classification (or learning) 
is that classes appeared in the test data must have appeared in training, called the \textit{closed-world} assumption ~\cite{fei2016breaking,ChenLiu2016}. Although this assumption holds in many applications, it is violated in many others, especially in dynamic or open environments. For example, in social media, a classifier built with past topics or classes may not be effective in classifying future data because new topics appear constantly in social media ~\cite{FeiLiu2016}. This is clearly true in other domains too, e.g., self-driving cars, where new objects may appear in the scene all the time. 


Ideally, in the text domain, the classifier should classify incoming documents to the right existing classes used in training and also detect those documents that don't belong to any of the existing classes. This problem is called~\textit{open world classification} or~\textit{open classification}~\cite{fei2016breaking}. Such a classifier is aware \textit{what it does and does not know}.
This paper proposes a novel technique to solve this problem.
	
	\textbf{Problem Definition}: Given the training data $D = \{(\mathbf{x}_1, y_1), (\mathbf{x}_2, y_2), \dots, (\mathbf{x}_n, y_n)\}$, where $\mathbf{x}_i$ is the $i$-th document,
	and $y_i \in \{l_1, l_2,\dots, l_m\} = \mathcal{Y}$ is $\mathbf{x}_i$'s class label, we want to build a model $f(\mathbf{x})$ that can classify each test instance $\mathbf{x}$ to one of the $m$ training or \textit{seen} classes in $\mathcal{Y}$ or reject it to indicate that it does not belong to any of the $m$ training or seen classes, i.e., \textit{unseen}. In other words, we want to build a $(m+1)$-class classifier $f(\mathbf{x})$ with the classes $\mathcal C =\{l_1, l_2,\dots, l_m, \textit{rejection}\}$. 
	
	
	There are some prior approaches for open classification. One-class SVM \cite{scholkopf2001estimating,tax2004support} is the earliest approach. However, as no negative training data is used, one-class classifiers work poorly.
	~\citet{fei2016breaking} proposed a Center-Based Similarity (CBS) space learning method~\cite{FeiLiu2015}. This method first computes a center for each class and transforms each document to a vector of similarities to the center. A binary classifier is then built using the transformed data for each class. The decision surface is like a ``ball'' encircling each class. Everything outside the ball is considered not belonging to the class. Our proposed method outperforms this method remarkably.~\citet{FeiLiu2016} further added the capability of incrementally or cumulatively learning new classes, which connects this work to~\textit{lifelong learning}~\cite{ChenLiu2016} because without the ability to identify novel or new things and learn them, a system will never be able to learn by itself continually. 
		
	In computer vision, \citet{scheirer2013toward}  studied the problem of recognizing unseen images that the system was not trained for by reducing \textit{open space risk}. The basic idea is that a classifier should not cover too much open space where there are few or no training data. They proposed to reduce the half-space of a binary SVM classifier with a positive region bounded by two parallel hyperplanes. 
	Similar works were also done in a probability setting by  \citet{scheirer2014probability} and 
	\citet{jain2014multi}. Both approaches use probability threshold, but choosing thresholds need prior knowledge, which is a weakness of the methods. \citet{dalvi2013exploratory} proposed a multi-class semi-supervised method based on the EM algorithm.
	It has been shown that these methods are poorer than the method in \cite{fei2016breaking}.
	
	The work closest to ours is that in \cite{bendale2016towards}, which leverages an algorithm called \emph{OpenMax} to add the rejection capability by utilizing the logits that are trained via closed-world softmax function. One weak assumption of OpenMax is that examples with equally likely logits are more likely from the unseen or rejection class, which can be examples that are hard to classify. Another weakness is that it requires validation examples from the unseen/rejection class to tune the hyperparameters. Our method doesn't make these weak assumptions and performs markedly better.  
	\label{sec:pre}
	
	\begin{figure}[t]
		\centering
		\includegraphics[width=0.48\textwidth]{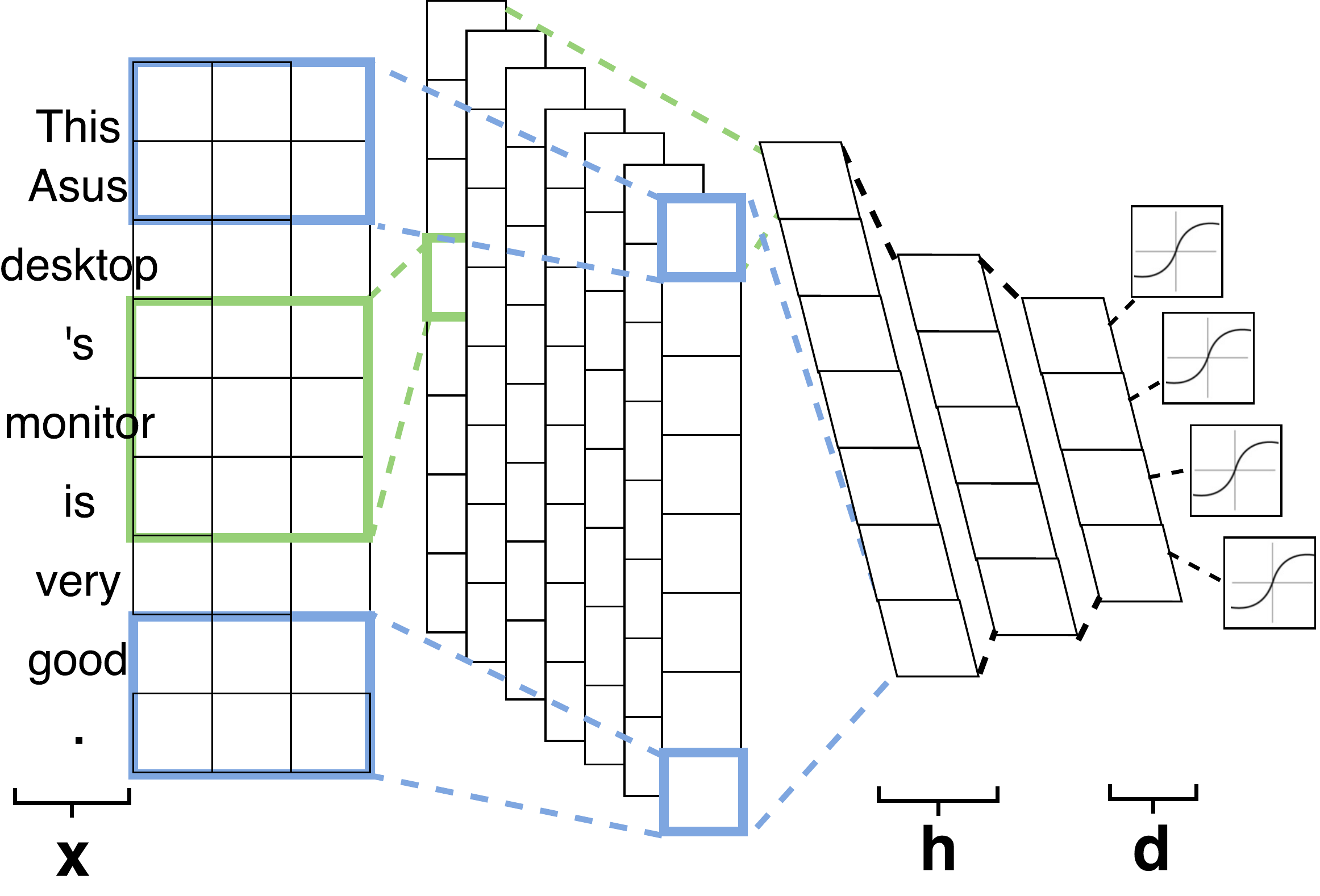}
		\caption{Overall Network of DOC}
		\label{fig:cnn}
	\end{figure}

	Our proposed method, called DOC (\textit{Deep Open Classification}), uses deep learning \cite{Goodfellow-et-al-2016,kim2014convolutional}. Unlike traditional classifiers, DOC builds a multi-class classifier with a 1-vs-rest final layer of sigmoids rather than softmax to reduce the open space risk. It reduces the open space risk further for rejection by tightening the decision boundaries of sigmoid functions with Gaussian fitting. Experimental results show that DOC dramatically outperforms state-of-the-art existing approaches from both text classification and image classification domains.

	\section{The Proposed DOC Architecture}
	DOC uses CNN \cite{collobert2011natural, kim2014convolutional} as its base and augments it with a 1-vs-rest final sigmoid layer and Gaussian fitting for classification. Note: other existing deep models like RNN \cite{williams1989learning,schuster1997bidirectional} and LSTM \cite{hochreiter1997long, gers2002learning} can also be adopted as the base. Similar to RNN, CNN also works on embedded sequential data (using 1D convolution on text instead of 2D convolution on images). We choose CNN because OpenMax uses CNN and CNN performs well on text \cite{kim2014convolutional}, which enables a fairer comparison with OpenMax.

	\subsection{CNN and Feed Forward Layers of DOC}
	
	The proposed DOC system 
	(given in Fig.~\ref{fig:cnn}) is a variant of the CNN architecture \cite{collobert2011natural} for text classification \cite{kim2014convolutional}\footnote{\url{https://github.com/alexander-rakhlin/CNN-for-Sentence-Classification-in-Keras}}. The first layer embeds words in document $\mathbf{x}$ into dense vectors. The second layer performs convolution over dense vectors using different filters of varied sizes (see Sec.~3.4). Next, the max-over-time pooling layer selects the maximum values from the results of the convolution layer to form a $k$-dimension feature vector $\bm h$.
	Then we reduce $\bm h$ to a $m$-dimension vector $\bm d=d_{1:m}$ ($m$ is the number of training/seen classes) via 2 fully connected layers and one intermediate ReLU activation layer: 
	\begin{equation}
		\begin{split}
			\bm d=\bm W' (\text{ReLU}(\bm W \bm h+ \bm b)) + \bm b',
		\end{split}
	\end{equation}
	where $\bm W \in \mathbb R^{r\times k}$, $\bm b \in \mathbb R^r$, $\bm W' \in \mathbb R^{m\times r}$, and $\bm b' \in \mathbb R^m$ are trainable weights; $r$ is the output dimension of the first fully connected layer. The output layer of DOC is a 1-vs-rest layer applied to $d_{1:m}$, which allows rejection. We describe it next.

	\subsection{1-vs-Rest Layer of DOC}
	Traditional multi-class classifiers \cite{Goodfellow-et-al-2016,bendale2016towards} typically use softmax as the final output layer, which does not have the rejection capability since the probability of prediction for each class is normalized across all training/seen classes. 
	Instead, we build a 1-vs-rest layer containing $m$ sigmoid functions for $m$ seen classes. For the $i$-th sigmoid function corresponding to class $l_i$, DOC takes all examples with $y=l_i$ as positive examples and all the rest examples $y \ne l_i$ as negative examples. 
	
	The model is trained with the objective of summation of all log loss of the $m$ sigmoid functions on the training data $D$.
	\begin{equation}
	\begin{split}
	    \text{Loss} = \sum_{i = 1}^{m}\sum_{j = 1}^{n} -\mathbb{I}(y_j = l_i) \log p(y_j = l_i)\\ -\mathbb{I}(y_j \neq l_i) \log (1- p(y_j = l_i)),
	\end{split}
	\end{equation}
	where $\mathbb{I}$ is the indicator function and $p(y_j = l_i) = \text{Sigmoid}(d_{j,i})$ is the probability output from $i$th sigmoid function on the $j$th document's $i$th-dimension of $\bm d$.
	
	During testing, we reinterpret the prediction of $m$ sigmoid functions to allow rejection, as shown in Eq. \ref{eq:rej}. For the $i$-th sigmoid function, we check if the predicted probability $\text{Sigmoid}(d_i)$ is less than a threshold $t_i$ belonging to class $l_i$. If all predicted probabilities are less than their corresponding thresholds for an example, the example is rejected; otherwise, its predicted class is the one with the highest probability. Formally, we have
	
	\begin{equation} \label{eq:rej}
		\resizebox{0.42\textwidth}{!}{
			$\hat{y} = \left\{
			\begin{array}{l}
			\textit{reject},\text{ if } \text{Sigmoid}(d_i)<t_i, \forall l_i \in \mathcal Y ;\\
			\argmax_{l_i \in \mathcal{Y}} \text{Sigmoid}(d_i),\text{ otherwise.}
			\end{array} \right.
			$ }
	\end{equation}
	
	Note that although multi-label classification \cite{huang2013multi, zhang2006multilabel, tsoumakas2006multi} may also leverage multiple sigmoid functions, Eq. \ref{eq:rej} forbids multiple predicted labels for the same example, which is allowed in multi-label classification. DOC is also related to multi-task learning \cite{ huang2013multi, caruana1998multitask}, where each label $l_i$ is related to a 1-vs-rest binary classification task with shared representations from CNN and fully connected layers. However, Eq. \ref{eq:rej} performs classification and rejection based on the outputs of these binary classification tasks.
	
	{\bf Comparison with OpenMax}: OpenMax builds on the traditional closed-world multi-class classifier (softmax layer). It reduces the open space for each seen class, which is weak for rejecting unseen classes. DOC's 1-vs-rest sigmoid layer provides a reasonable representation of all other classes (the rest of seen classes and unseen classes), and enables the 1 class forms a good boundary. Sec.~3.5 shows that this basic DOC is already much better than OpenMax. Below, we improve DOC further by tightening the decision boundaries more.

	\subsection{Reducing Open Space Risk Further}
	\begin{figure}[t] 
		\centering
		\includegraphics[width=0.48\textwidth]{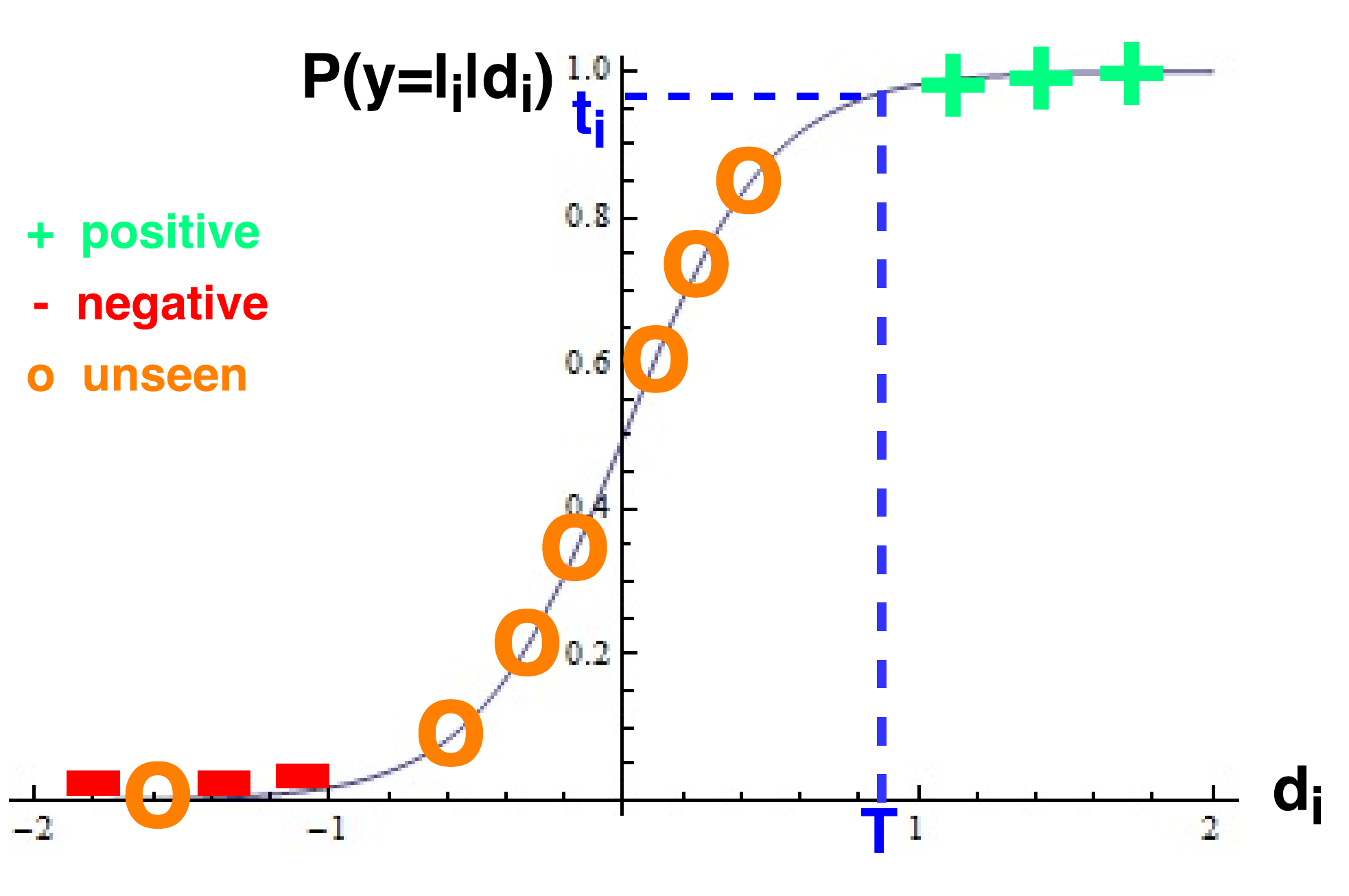}
		\caption{Open space risk of sigmoid function and desired decision boundary $d_i = T$ and probability threshold $t_i$.}
		\label{fig:sigmoid}
	\end{figure}
	
	Sigmoid function usually uses the default probability threshold of $t_i$ = 0.5 for classification of each class $i$. But this threshold does not consider potential open space risks from unseen (rejection) class data. We can improve the boundary by increasing $t_i$. We use Fig.~\ref{fig:sigmoid} to illustrate. 
	The x-axis represents $d_i$ and y-axis is the predicted probability $p(y=l_i|d_i)$. The sigmoid function tries to push positive examples (belonging to the $i$-th class) and negative examples (belonging to the other seen classes) away from the y-axis via a high gain around $d_i=0$, which serves as the default decision boundary for $d_i$ with $t_i = 0.5$. 
	As demonstrated by those 3 circles on the right-hand side of the y-axis, during testing, unseen class examples (circles) can easily fill in the gap between the y-axis and those dense positive (+) examples, which may reduce the recall of rejection and the precision of the $i$-th seen class prediction. Obviously, a better decision boundary is at $d_i=T$, where the decision boundary more closely ``wrap'' those dense positive examples with the probability threshold $t_i \gg 0.5$ . 
	
	To obtain a better $t_i$ for each seen class $i$-th, we use the idea of outlier detection in statistics: 
	
\begin{enumerate}

	\item Assume the predicted probabilities $p(y=l_i| \mathbf{x}_j, y_j = l_i )$ of all training data of each class $i$ follow one half of the Gaussian distribution (with mean $\mu_i=1$), e.g., the three positive points in Fig. 2 projected to the y-axis (we don't need $d_i$). 
    We then artificially create the other half of the Gaussian distributed points ($\ge 1$): for each existing point $p(y=l_i| \mathbf{x}_j, y_j = l_i )$, we create a mirror point $1+ (1-p(y=l_i|\mathbf{x}_j, y_j = l_i)$ (not a probability) mirrored on the mean of 1. 
	
    \item Estimate the standard deviation $\sigma_i$ using both the existing points and the created points. 
    
    \item In statistics, if a value/point is a certain number ($\alpha$) of standard deviations away from the mean, it is considered an outlier. We thus set the probability threshold $t_i=\max (0.5, 1-\alpha \sigma_i)$. The commonly used number for $\alpha$ is 3, which also works well in our experiments. 

\end{enumerate}

	Note that due to Gaussian fitting, different class $l_i$ can have a different classification threshold $t_i$. 
		
	


	
	\section{Experimental Evaluation}
	\label{sec:exp}
	
	\subsection{Datasets}
	We perform evaluation using two publicly available datasets, which are exactly the same datasets used in \cite{fei2016breaking}. 
	
	(1) \textbf{20 Newsgroups}\footnote{\url{http://qwone.com/ ~jason/20Newsgroups/}}~\cite{20news}: The 20 newsgroups data set contains 20 non-overlapping classes. Each class has about 1000 documents. 
	
	(2) \textbf{50-class reviews}~\cite{chen2014mining}: The dataset has Amazon reviews of 50 classes of products. Each class has 1000 reviews. Although product reviews are used, we do not do sentiment classification. We still perform topic-based classification. That is, given a review, the system decides what class of product the review is about. 
	
	For every dataset, we keep a 20000 frequent word vocabulary. Each document is fixed to 2000-word length (cutting or padding when necessary).
	
	\subsection{Test Settings and Evaluation Metrics}
	For a fair comparison, we use exactly the same settings as in \cite{fei2016breaking}. For each class in each dataset, we randomly sampled 60\% of documents for training, 10\% for validation and 30\% for testing. \citet{fei2016breaking} did not use a validation set, but the test data is the same 30\%. We use the validation set to avoid overfitting.
	For open-world evaluation, we hold out some classes (as unseen) in training and mix them back during testing.
	We vary the number of training classes and use 25\%, 50\%, 75\%, or 100\% classes for training and all classes for testing. Here using 100\% classes for training is the same as the traditional closed-world classification. Taking 20 newsgroups as an example, for 25\% classes, we use 5 classes (we randomly choose 5 classes from 20 classes for 10 times and average the results, as in \cite{fei2016breaking}) for training and all 20 classes for testing (15 classes are unseen in training). We use macro $F_1$-score over $5+1$ classes (1 for rejection) for evaluation. Please note that examples from unseen classes are dropped in the validation set.
	
	\subsection{Baselines}
	We compare DOC with two state-of-the-art methods published in 2016 and one DOC variant. 
	
	
	\textbf{cbsSVM}: This is the latest method published in NLP \cite{fei2016breaking}. It uses SVM to build 1-vs-rest CBS classifiers for multiclass text classification with rejection option. 
	The results of this system are taken from \cite{fei2016breaking}.
	
	\textbf{OpenMax}: This is the latest method from computer vision \cite{bendale2016towards}. Since it is a CNN-based method for image classification, we adapt it for text classification by using CNN with a softmax output layer, and adopt the OpenMax layer\footnote{\url{https://github.com/abhijitbendale/OSDN}} for open text classification. When all classes are seen (100\%), the result from softmax is reported since OpenMax layer always performs rejection. We use default hyperparameter values of OpenMax (Weibull tail size is set to 20). 
	
	\textbf{DOC$(t = 0.5)$}: This is the basic DOC $(t = 0.5)$. Gaussian fitting isn't used to choose each $t_i$.
	
	Note that \cite{fei2016breaking} compared with several other baselines. We don't compare with them as it was shown that cbsSVM was superior. 
	
	\subsection{Hyperparameter Setting}	
	We use word vectors pre-trained from Google News\footnote{\url{https://code.google.com/archive/p/word2vec/}} (3 million words and 300 dimensions). For the CNN layers, 3 filter sizes are used $[3, 4, 5]$. For each filter size, 150 filters are applied. The dimension $r$ of the first fully connected layer is 250.

	\begin{table}[!]
		\centering
		\caption{Macro-$F_1$ scores for 20 newsgroups}\label{tab:20news}  
		\resizebox{0.5\textwidth}{!}{
		\begin{tabular}{c|c|c|c|c}                      
			{\bf \% of seen classes}& {\bf 25\%}&{\bf 50\%}& {\bf 75\%}& {\bf 100\%}  \\\hline\hline
			cbsSVM & 59.3 & 70.1 & 72.0 & 85.2 \\\hline
			OpenMax &35.7 &59.9 &76.2 & 91.9\\\hline
			DOC ($t=0.5$) &75.9& 84.0& 87.4& 92.6 \\\hline
			DOC &82.3& 85.2& 86.2& 92.6\\\hline                
		\end{tabular} 
		}
	\end{table}
	\begin{table}[!]
	    \centering
		\caption{Macro-$F_1$ scores for 50-class reviews}\label{tab:50products}
	    \resizebox{0.5\textwidth}{!}{
		\begin{tabular}{c|c|c|c|c}                      
			{\bf \% of seen classes}& {\bf 25\%}&{\bf 50\%}& {\bf 75\%}& {\bf 100\%}  \\\hline\hline
			cbsSVM & 55.7 & 61.5 & 58.6 & 63.4 \\\hline
			OpenMax &41.6& 57.0 & 64.2& 69.2\\\hline
			DOC ($t=0.5$) &51.1& 63.6& 66.2& 69.8 \\\hline
			DOC &61.2& 64.8& 66.6& 69.8\\\hline			
		\end{tabular}
		}
	\end{table}
	
	\subsection{Result Analysis}
	The results of 20 newsgroups and 50-class reviews are given in Tables \ref{tab:20news} and \ref{tab:50products}, respectively. From the tables, we can make the following observations: 
	
    \begin{enumerate}
    
	\item DOC is markedly better than OpenMax and cbsSVM in macro-$F_1$ scores for both datasets in the 25\%, 50\%, and 75\% settings. For the 25\% and 50\% settings (most test examples are from unseen classes), DOC is dramatically better. Even for 100\% of traditional closed-world classification, it is consistently better too. DOC($t=0.5$) is better too. 
	\item For the 25\% and 50\% settings, DOC is also markedly better than DOC($t=0.5$), which shows that Gaussian fitting finds a better probability threshold than $t=0.5$ when many unseen classes are present. 
	In the 75\% setting (most test examples are from seen classes), DOC($t=0.5$) is slightly better for 20 newsgroups but worse for 50-class reviews. DOC sacrifices some recall of seen class examples for better precision, while $t=0.5$ sacrifices the precision of seen classes for better recall. DOC($t=0.5$) is also worse than cbsSVM for 25\% setting for 50-class reviews. It is thus not as robust as DOC.
	
	\item For the 25\% and 50\% settings, cbsSVM is also markedly better than OpenMax.  
    \end{enumerate}
	
	\section{Conclusion}
	
	This paper proposed a novel deep learning based method, called DOC, for open text classification. Using the same text datasets and experiment settings, we showed that DOC performs dramatically better than the state-of-the-art methods from both the text and image classification domains. We also believe that DOC is applicable to images. 
	
	In our future work, we plan to improve the cumulative or incremental learning method in~\cite{FeiLiu2016} to learn new classes without training on all past and new classes of data from scratch. This will enable the system to learn by self to achieve continual or lifelong learning~\cite{ChenLiu2016}. We also plan to improve model performance during testing~\cite{ShuXuLiu2017}.  

\section*{Acknowledgments} 
This work was supported in part by grants from National Science Foundation (NSF) under grant no. IIS-1407927 and IIS-1650900. 

\bibliography{emnlp2017}
\bibliographystyle{emnlp_natbib}

\end{document}